\documentclass{bmvc2k}

\title{Analysis of Training Object Detection Models with Synthetic Data}

\addauthor{Bram Vanherle}{bram.vanherle@uhasselt.be}{1}
\addauthor{Steven Moonen}{steven.moonen@uhasselt.be}{1}
\addauthor{Frank Van Reeth}{frank.vanreeth@uhasselt.be}{1}
\addauthor{Nick Michiels}{nick.michiels@uhasselt.be}{1}

\addinstitution{
  Hasselt University - tUL \\ Flanders Make, \\
  Expertise Centre for Digital Media
}

\runninghead{Vanherle et al.}{Analysis of Object Detection with Synthetic Data}

\def\etal{\emph{et al}\bmvaOneDot}

\begin{document}

\maketitle

\begin{abstract}
Recently, the use of synthetic training data has been on the rise as it offers correctly labelled datasets at a lower cost. The downside of this technique is that the so-called domain gap between the real target images and synthetic training data leads to a decrease in performance. In this paper, we attempt to provide a holistic overview of how to use synthetic data for object detection. We analyse aspects of generating the data as well as techniques used to train the models. We do so by devising a number of experiments, training models on the Dataset of Industrial Metal Objects (DIMO)~\cite{DIMO}. This dataset contains both real and synthetic images. The synthetic part has different subsets that are either exact synthetic copies of the real data or are copies with certain aspects randomised. This allows us to analyse what types of variation are good for synthetic training data and which aspects should be modelled to closely match the target data. Furthermore, we investigate what types of training techniques are beneficial towards generalisation to real data, and how to use them. Additionally, we analyse how real images can be leveraged when training on synthetic images. All these experiments are validated on real data and benchmarked to models trained on real data. The results offer a number of interesting takeaways that can serve as basic guidelines for using synthetic data for object detection. Code to reproduce results is available at \url{https://github.com/EDM-Research/DIMO_ObjectDetection}.

\end{abstract}

\section{Introduction}
Deep learning and its applications have advanced tremendously over the last couple of years. These powerful machine learning models require a large amount of labelled training data however. The more complex and the better these models get, the more training data they require. But good training data is not easy to come by. Manually creating photographs and labeling them is a slow and costly process. Additionally, humans are prone to introducing errors and bias in datasets~\cite{dataset_bias, dataset_error}, which is bad for model performance. Furthermore, some forms of annotations are very difficult for a human to create, such as depth maps, segmentation maps or object poses. \\
Due to these problems with datasets created by humans, synthetic training data has become more popular over recent years. With modern rendering technology it is easy to render thousands of images fairly quickly and at a low cost when 3D models are provided. Since the 3D composition of the depicted scene is known, the accompanying labels for the machine learning task can easily be generated. Additionally, these labels are pixel correct and the dataset contains less bias, since a computer is way better at randomising than a human. There is however a big disadvantage to using synthetic data. Although looking very realistic, there still is a difference in appearance between real and rendered images. This causes a model that is trained on synthetic images, to perform worse on real images. This phenomenon is called the domain gap~\cite{domain_randomization_tobin} and it hinders synthetic training data from being widely adopted. \\
Object Detection is one of the most prominent fields of computer vision. This is due to the fact that it has many applications and is often the first step in vision pipelines for more complex tasks. Some of these applications include robot control~\cite{robot_control}, product inspection~\cite{product_inspection}, surveillance~\cite{surveillance} and many more. If a company wants to apply deep learning to their specific tasks, they need high quality training data that is specific to them. Manually labelled data is often too expensive or sometimes even too difficult to come by, especially so for smaller companies. These companies sometimes resolve to using synthetic data to train their models, often with unsatisfactory results, due to the aforementioned problems with synthetic data. While synthetic data is cheaper then manually created data, it is not for free. When rendering thousands of images, costs can accumulate to a large number as well. \\
In this paper we offer a number of insights on how to generate and how to use synthetic training data. The goal is to generate knowledge on how to create training data that offers good performance on real images whilst keeping the total cost of rendering as low as possible. When using this synthetic data we use only basic deep learning mechanisms that are available in most toolkits. We deliberately stray away from more complex methods of domain adaptation and generalisation to make our findings as widely applicable as possible. To provide these insights, we perform a number of experiments using the Dataset of Industrial Metal Objects (DIMO)~\cite{DIMO}. This dataset contains a set of real images, exact synthetically rendered copies of those real images and sets of synthetic images with variations in different aspects. This unique dataset allows us to study the exact impact of those variations towards the generalisation on real images. However, data alone is only half the picture. Additionally, we study the impact of a number of deep learning techniques towards the generalisation on real test sets. The impact is measured by training a number of object detection models on different datasets and configurations while measuring the performance on a real test set.

\section{Related Work}
Models trained on synthetic data often suffer from a decrease in performance on real data. This is due to the domain gap, a term introduced by Tobin \etal~\cite{domain_randomization_tobin}. They argue that it is impossible to perfectly simulate all aspects of a camera and that there will always be a difference between synthetic training data and real test data. They solve this for the task of object localisation by using domain randomisation. This technique randomises as many aspects of the rendering as possible as opposed to trying to accurately simulate the data. Trembley \etal~\cite{domain_randomization_tremblay} applied this technique for object detection. Their domain randomised car detection dataset leads to great performance on the KITTI dataset~\cite{kitti}, even better than the Virtual KITTI dataset~\cite{virtual_kitti} that was modelled to be similar. This has shown that randomisation can be a substitute for realism.\\
A different approach to randomisation is attempting to make the datasets as realistic as possible. Movshovitz-Attias \etal~\cite{photorealism_movshovitz} investigated how useful photorealism is and what parameters are the most important, for the task of viewpoint estimation. They show that a more complex rendering process is beneficial and that adding synthetic images to a real dataset offers a boost in performance. Additionally, they conclude that randomising lighting parameters leads to better generalisation. Hodan \etal~\cite{photorealism_hodan} developed a method for generating object detection datasets using physically based rendering (PBR). They show that models trained on PBR datasets perform better than ones trained on datasets created by simpler rendering techniques and that increasing the quality of the PBR leads to better models. Additionally, they show that taking into account the context (gist, geometric, semantic, and illumination contextual aspects) in which the object will be placed improves the performance of the trained network. \\
There are other techniques to improve performance as well. Hinterstoisser \etal~\cite{pretrained_features} use transfer learning to improve generalisation of models trained on synthetic data. They initialise a network with weights trained on real data and freeze the layers of the feature detector during training. Using this technique, they train a model on a simple synthetic dataset and manage to get performance close to that of a model trained on real data. Nowruzi \etal~\cite{real_data} investigated the use of real images when training on synthetic data. They show that adding a small amount of real images can be useful and that fine-tuning is better than mixing the data. \\
In our research we use the DIMO dataset. This dataset contains a real dataset, an exact synthetic copy and a number of different variations of the synthetic copy. We can thus investigate what variations are beneficial for generalisation but also research if it is useful to put effort in copying some aspects of the target dataset. This allows us to offer some unique insights that other papers have not yet investigated. Additionally, we attempt to offer a holistic analysis, investigating a number of important concepts for synthetic training data.

\section{Experimental Setup} \label{sec:setup}

\subsection{The Dataset} \label{sec:dataset}
\begin{figure}
    \centering
     \includegraphics[width=0.19\textwidth]{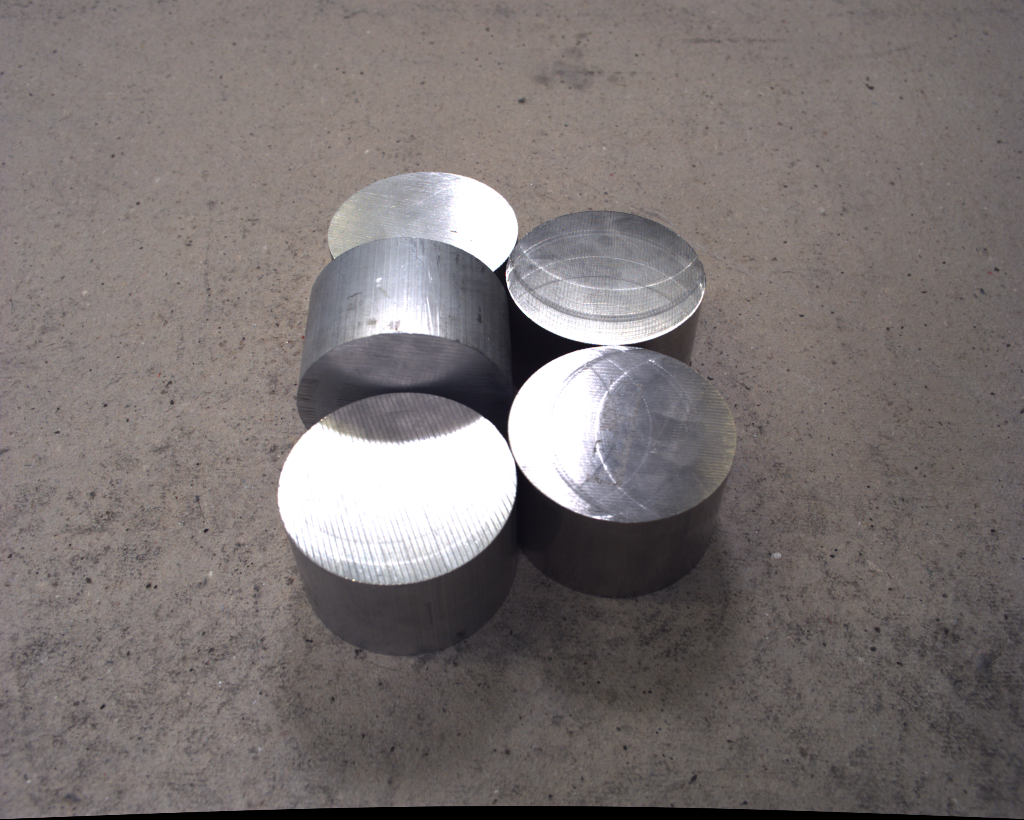}
     \includegraphics[width=0.19\textwidth]{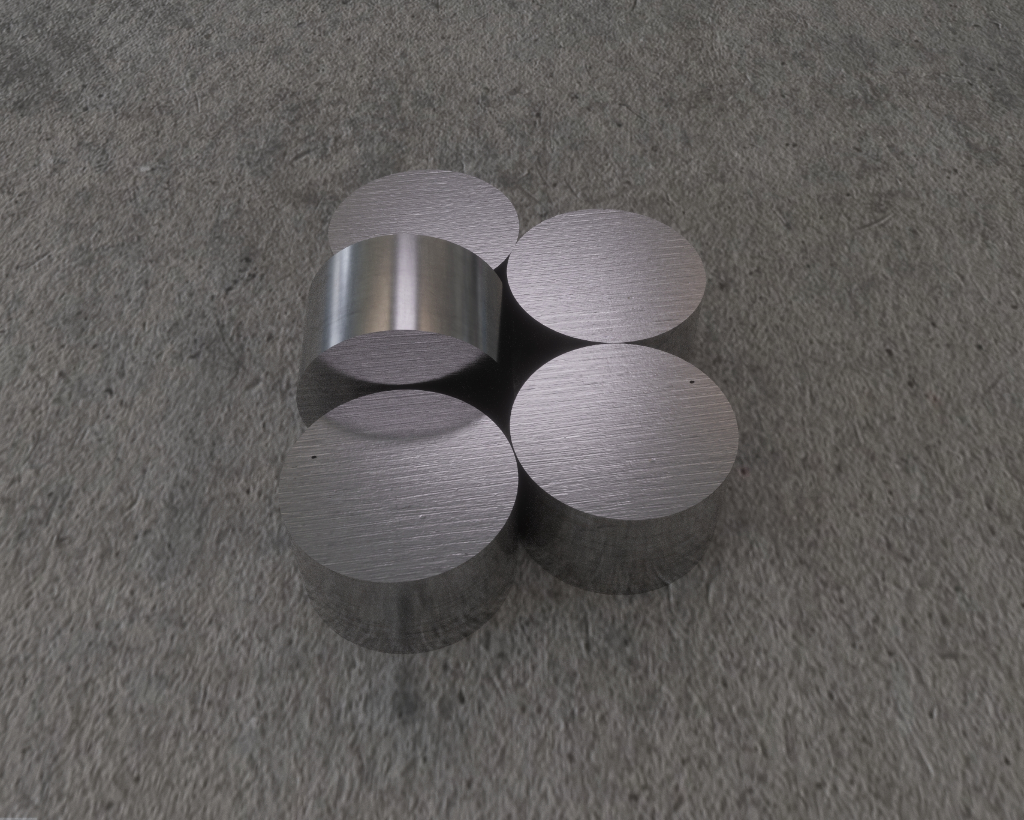}
     \includegraphics[width=0.19\textwidth]{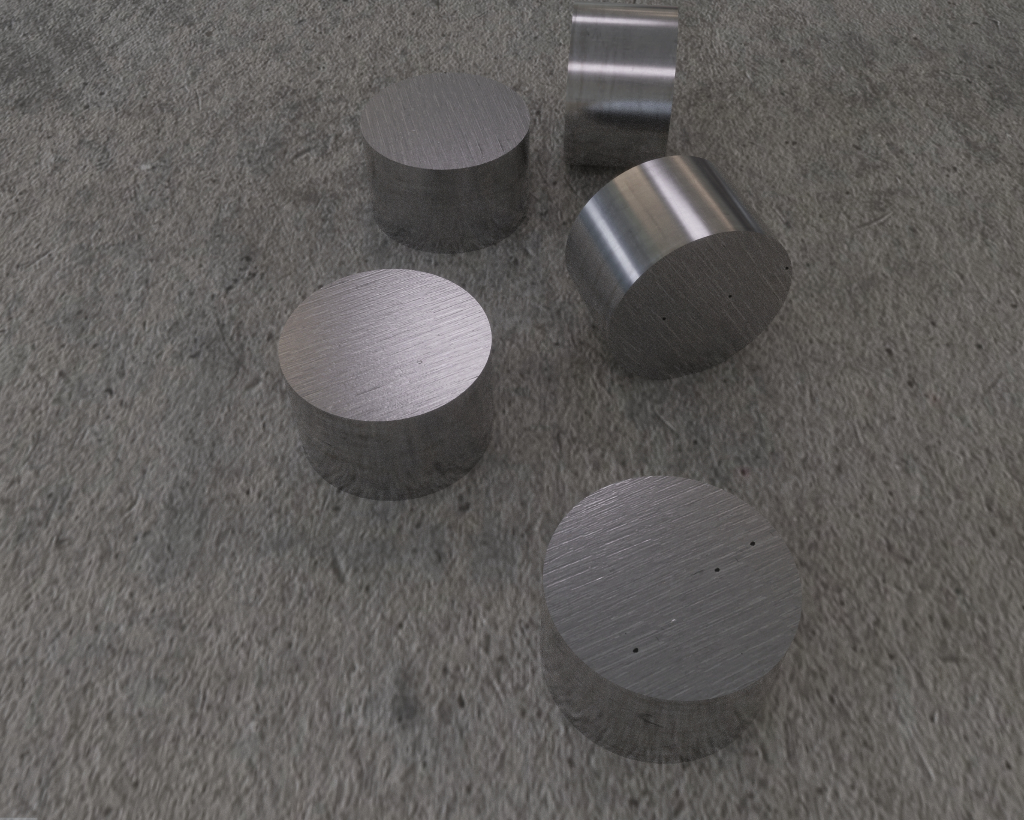}
     \includegraphics[width=0.19\textwidth]{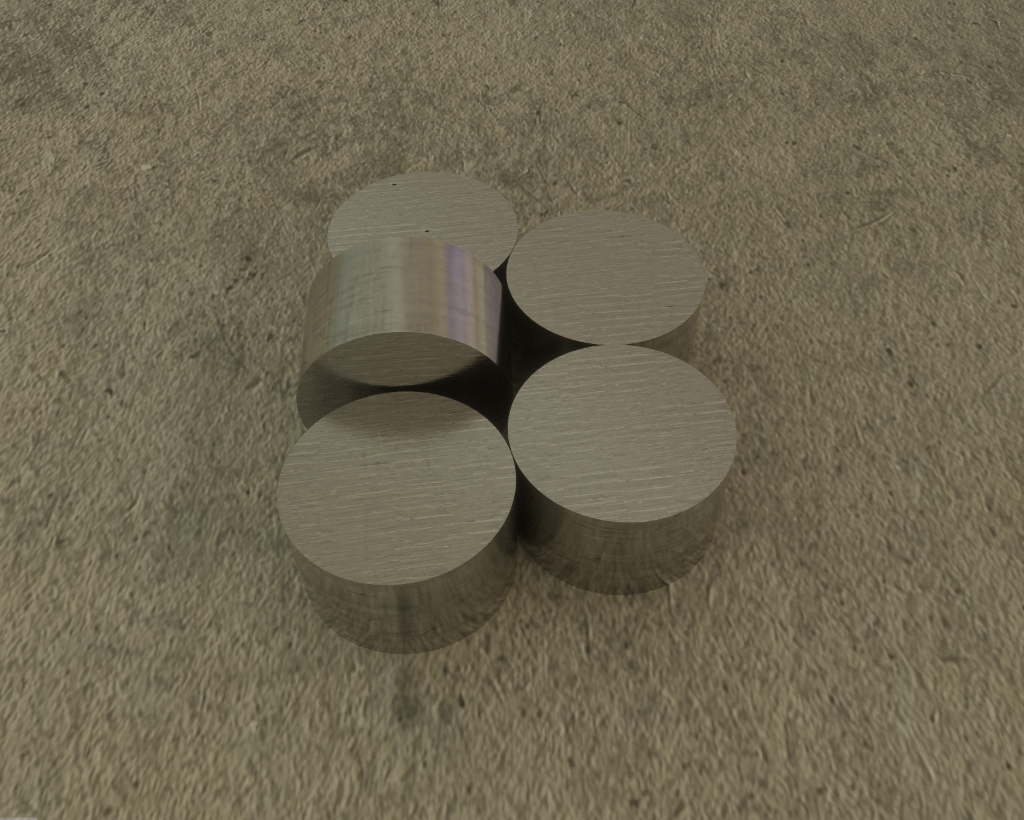}
     \includegraphics[width=0.19\textwidth]{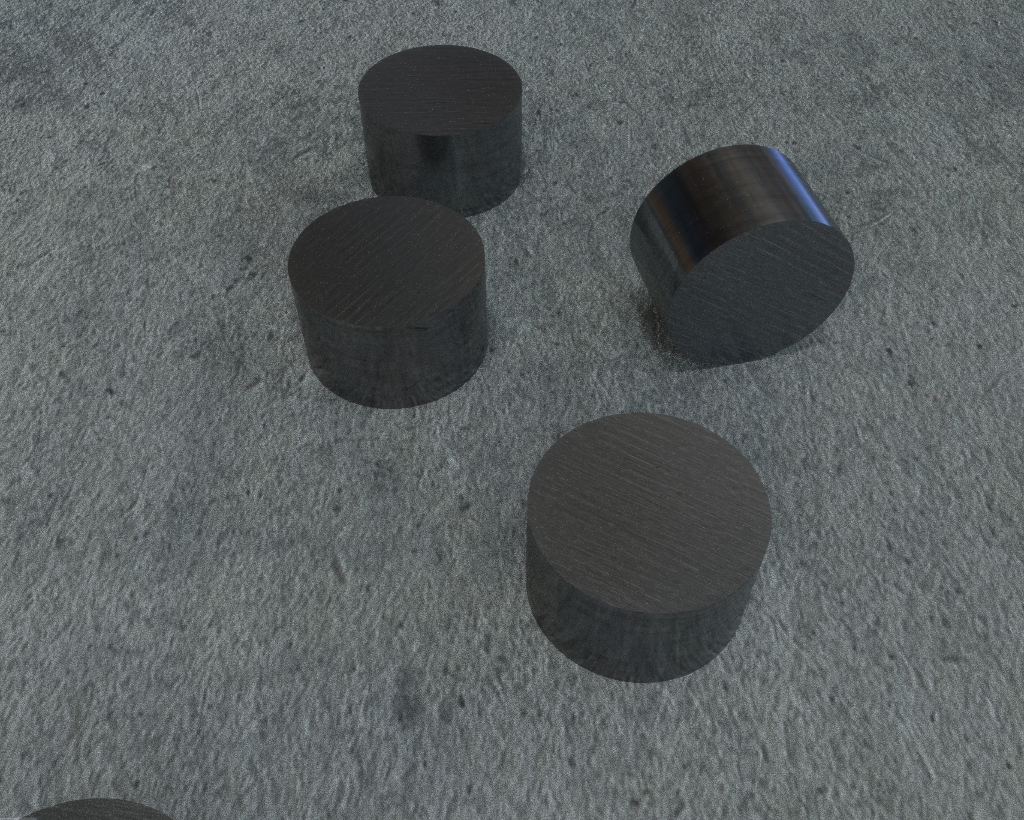}
    \caption{Examples of an scene in the DIMO dataset to illustrate the different variations. From left to right: Real image, synthetic copy, randomised poses, randomised lighting, both randomised.}
    \label{fig:dimo_sample}
\end{figure}

In our experiments we use five subsets of the DIMO datasets. The first subset contains real RGB images, captured with a JAI GO-5000 camera (subset denoted as \textit{real}). Additionally, we used the four synthetic datasets that have two types of variations. The first is an exact digital twin dataset, where both the poses of the objects as well as the light of the environment match with the real images (\textit{synth}). The second and third are datasets for which either the poses (\textit{synth, rand pose}) or the lighting conditions (\textit{synth, rand light}) are randomised, with the non-randomised component matching the real images. The fourth and last dataset of DIMO that we use in our experiments varies both the poses and the lighting conditions (\textit{synth, rand all}). The poses of the objects for the real images were manually set in representative and interesting positions. These poses were then manually annotated and used in the generation of the synthetic data. The randomization of the poses in the synthetic data is done by spawning the objects in a uniform random location 30 cm above the carrier. Subsequently, a physics engine simulates the dropping of the objects on the carrier, until they advance into a stable state. This provides a variation in the distribution of poses between the synthetic and real datasets. The light variations were created by iterating over a list of 31 environment maps of indoor scenes. The real light was captured with a HDR 360 image of the environment where the real images where recorded. More details on the creation of this dataset can be found in the original DIMO~\cite{DIMO} paper. Figure~\ref{fig:dimo_sample} shows an example scene from the DIMO dataset, and images from the four different synthetic datasets.

For the experiments in this paper, we use the first 150 scenes of the DIMO dataset for each of the subsets. Since there is more variation introduced in some datasets, they do not have an equal amount of images. The real and its synthetic twin have around 2k images each, the synthetic datasets with random light or random poses contain around 29k images each and the fully random synthetic dataset contains 78k images. We split each dataset in a training (90\%), validation (5\%) and test (5\%) set. Since the datsets are subdivided in scenes, we ensure that all images from a specific scene belong to the same set. The models are thus tested on unseen scenes.

\subsection{The Model} \label{sec:model}
Although recently transformers have surpassed convolutional neural networks in object detection performance~\cite{dino}, CNN's still remain the most popular type of neural networks for vision. This is especially true outside of research. Additionally, a lot of research is still being done in the field of CNN's, pushing their performance closer to that of transformers~\cite{convnext, yolov7}. We therefore opt to focus our analysis on CNN based feature detectors.\\
In our experiments we use the Mask R-CNN model~\cite{maskrcnn, matterport_maskrcnn}, a widely used object detection and instance segmentation model. It is a two stage architecture consisting of a convolutional feature detection network and detection heads. In this work ResNet101~\cite{resnet} is used as a feature detector. When transfer learning is used throughout this paper, the feature detector is initialised with weights trained on COCO~\cite{coco_dataset}. Unless mentioned otherwise, the layers of the feature detector are frozen when using transfer learning. In each experiment we train the model for 100 epochs with Stochastic Gradient Descent using a learning rate of $0.001$ and a momentum of $0.9$. A batch size of four is used and each epoch $1.000$ images are used to train the model. This is done to be able to consistently compare per-epoch model performance between models trained on datasets with different amounts of images. If data augmentation is used this is a combination of zero to two color modifying augmentations and zero to one translating augmentations. The color augmentations include: add, multiply, Gaussian blur, Gaussian noise, motion blur and grayscale; the translating augmentations include: rotation, translation, shear, scale, horizontal flip and perspective transform. More details are provided in the supplementary material.

\section{Results}
In this section we describe and execute a number of experiments using the setup described in Section~\ref{sec:setup}. To test the performance on the real target domain for each of these experiments, we test the trained models on an unseen set of real images. We compute the AP, AP\textsubscript{50} and AP\textsubscript{75} values as described for the COCO challenge~\cite{coco_dataset}. In this section we only report on the AP value as the other metrics follow the same trends. For completeness, the other metrics are provided in the supplementary material.

\subsection{Scene Composition}
\begin{figure}
    \centering
    \includegraphics[width=0.7\textwidth]{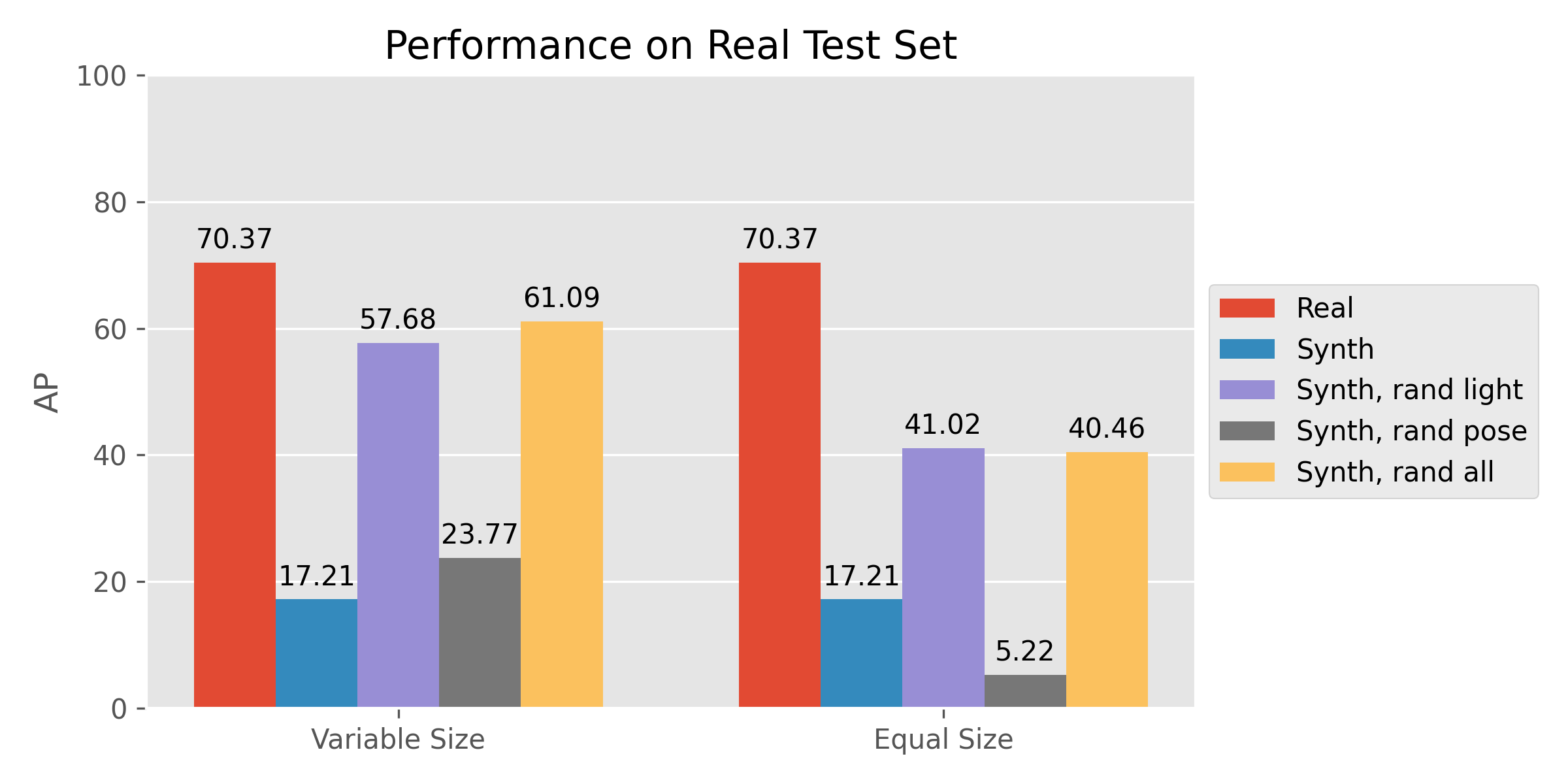}
    \caption{Results of the Mask R-CNN model trained on each of the five datasets, with variable and equal dataset sizes, on the real test dataset.}
    \label{fig:exp1}
\end{figure}

In this first experiment we attempt to determine whether it is important to create a synthetic dataset that closely matches the target data in terms of scene compositions and what type of variations are beneficial for generalisation. Should you use lighting conditions and object poses that are plausible for the real test dataset, or can you just make these parameters random? To find out, we trained a Mask R-CNN model from scratch on each of five datasets described in Section~\ref{sec:dataset}. Since the datasets have different amounts of images, we also repeated the experiment equalising the sizes of the datasets. Each of the datasets were reduced to the size of the real dataset, which is 1775 images. We sampled random images from the training set. The results are shown in Figure~\ref{fig:exp1}.

When looking at the results from the variable size experiment, we can see that the model trained on the real images performs the best on the real test set, this is as expected. The rendered copy of that dataset, with the same poses and lighting conditions, performs the worst and fails to generalise towards the real images. The model trained on the dataset with the same poses but with extra images under varying lighting conditions generalises reasonably well, falling 13 AP points under the model trained on real data. The model trained on the dataset with randomised poses and real lighting conditions performs way worse, only achieving 23.77 AP. This shows that variation in the form of lighting conditions is more beneficial toward generalisation on real data than variation in object poses. The model trained on the fully randomised dataset performs the best of all the synthetic datasets. This dataset is however way larger than the other datasets. \\
We therefore also compare the performance of the models trained on datasets of equal size. Here we see that the model trained on the dataset with randomised light and real poses performs the best of the synthetic datasets. It performs slightly better than the fully randomised dataset. Both of these models still outperform the non randomised synthetic dataset. The model trained on the dataset with real lighting and randomised poses performs the worst with only 5.22 AP. This leads us to conclude that there is a slight benefit to modelling object poses. We confirm our suspicion that varying lighting conditions help toward generalisation on real data and trying to model real light conditions hurts performance. We argue this is due to the fact that it is easier to make higher level features such as shapes and poses match between real and rendered images, it therefore makes sense to try and accurately simulate these features in the synthetic dataset. Lower level features -- such as color, lighting and texture -- are more difficult to accurately render in a synthetic dataset. We therefore believe it is better to try and randomise these features of the synthetic dataset, as this leads to better generalisation.

\subsection{Training Techniques}
In our previous experiment, we trained the model from scratch and did not augment our data. This is however not a realistic scenario. It has been shown that transfer learning and data augmentation help improve performance on real data, when trained on synthetic data~\cite{pretrained_features}. We therefore repeat the previous experiment, but now we include transfer learning and data augmentation to analyse their effects on generalisation. For this experiment, we are more focused on the effects of certain training techniques on the different datasets as opposed to comparing the datasets amongst each other. We therefore use all images of each dataset, the models have thus been trained on different sizes of datasets. We include one extra experiment where we restricted each dataset to have the same size as the real dataset, being 1.7k images.

\begin{figure}
    \centering
    \includegraphics[width=\textwidth]{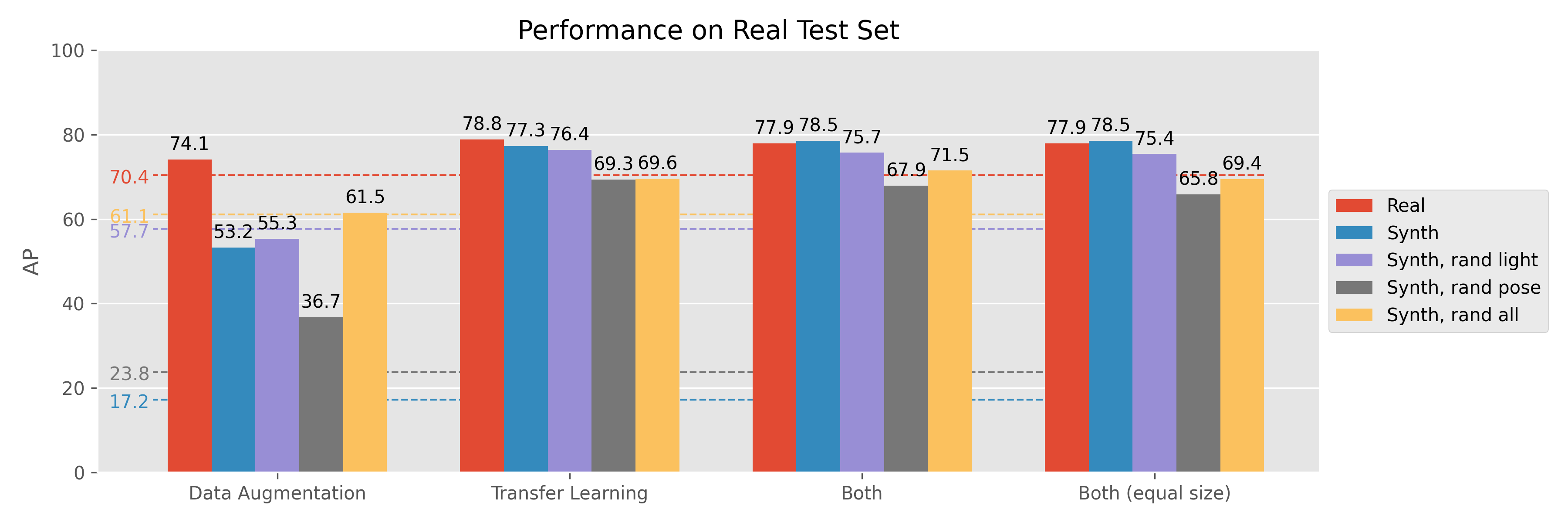}
    \caption{Results of training the model on each of the five datasets with data augmentation and/or transfer learning. The horizontal lines indicate the performance of the model trained without these techniques.}
    \label{fig:exp2}
\end{figure}

Figure~\ref{fig:exp2} shows the results of these experiments. The horizontal lines represent the performance of the models trained without any of these techniques. For the models trained with only data augmentation, we see that the model trained on real images gets a slight performance boost compared to the model without (3.7 AP). The models trained on the two datasets with randomised lighting experience only a small difference in accuracy and the model trained on the dataset with real poses and random lighting even suffers a small decrease in performance. The two models trained on the datasets with the real lighting conditions experience a large boost in performance. When training the models with transfer learning we observe a large boost in performance for all models. Interestingly, the worst performer of the previous experiment -- the synthetic copy dataset -- now becomes the best performing synthetic dataset, while having much less images than all the other datasets. When using both techniques we see very similar performance to when only transfer learning is used. Some models even suffer a slight decrease in accuracy. Finally, when we train the model on equally sized datasets using both training techniques, we notice only a very small decrease in performance. This is notable, since the dataset sizes decrease from 26k and 70k to only 1.7k.

\begin{figure}
    \centering
    \includegraphics[width=\textwidth]{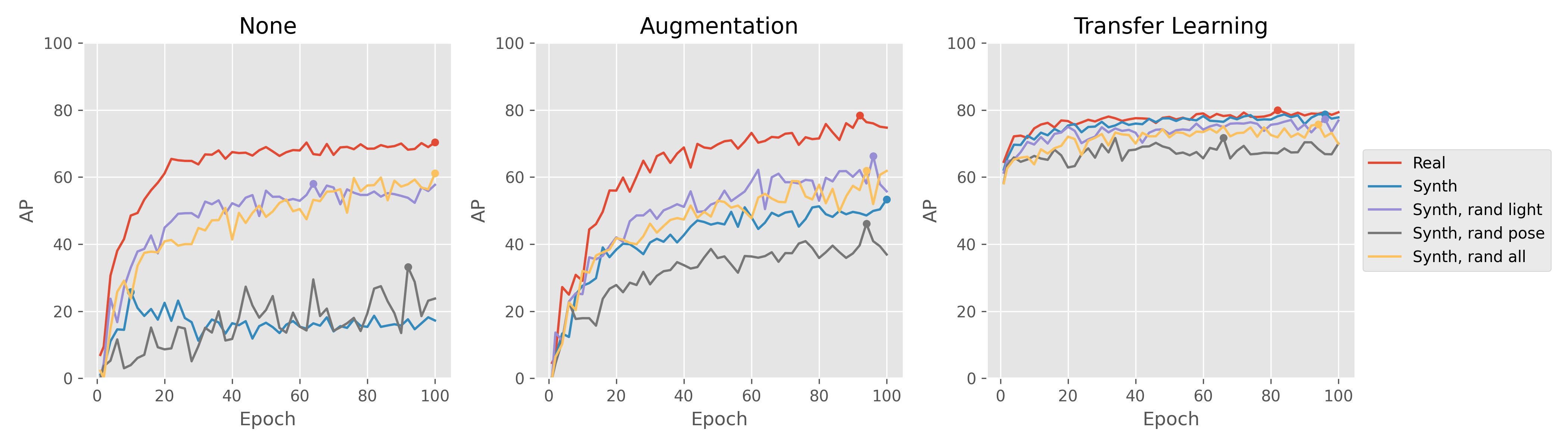}
    \caption{The evolution of the AP on the real test set for models trained without any techniques, with data augmentation and transfer learning.}
    \label{fig:exp2_epochs}
\end{figure}

To further analyse the impact of data augmentation and transfer learning, we compute the AP on the real test set for every two epochs. This allows us to investigate the evolution of the training process. Figure~\ref{fig:exp2_epochs} shows the training process for the model trained without any techniques and for the models trained with either data augmentation or transfer learning. When using no techniques we see that learning flattens of very quickly, especially for the datasets with no variation in lighting. When using data augmentation the model is able to learn for longer and at a better rate. This has a way larger impact on the datasets with no variation in lighting. When using transfer learning, the models for all of the datasets show good performance after one epoch, the learning flattens off very quickly however.\\
From these results we conclude that data augmentation and especially transfer learning help overcome the difference in low level features between real and synthetic data. Data augmentation does so by introducing more variation in these features, leading to a more robust model. Transfer learning achieves this by initialising the model with weights that are already capable of detecting low level features from real world images. Thus when using these techniques, it is beneficial to accurately simulate the poses and lighting conditions of the target domain in the synthetic dataset. When it is not possible to model the target domain, one should try to maximise variation in lighting conditions to achieve the best generalisation. Additionally, it is not necessary to use a large amount of images, even when using a randomised dataset. To confirm this, we trained models on a number of subsequently smaller subsets of the full random synthetic dataset. The results, shown in Table~\ref{tab:exp3}, indicate that adding more images only helps until a certain amount as we see a peak at 20k images. The differences in AP are not big, showing that only a few thousand images can already produce a decent model.

\begin{table}
    \begin{center}
    \begin{tabular}{|l|c|}
    \hline
    Image Count & AP \\
    \hline\hline
    1755    & 69.42 \\
    4387    & 72.23 \\
    8775    & 72.58 \\
    17550   & \textbf{73.01} \\
    35100   & 72.01 \\
    70200   & 71.52 \\
    \hline
    \end{tabular}
    \end{center}
    \caption{Results of training the model on different amounts of images sampled from the fully random dataset.}
    \label{tab:exp3}
\end{table}

\subsubsection{Transfer Learning}
So far we have shown that transfer learning is very helpful when training on synthetic data. Transfer learning can be done in multiple ways however. In our previous experiments we initialised the feature detector with weights trained on COCO, froze those layers and only trained the network heads. This forces the network to make predictions based on features learned from the COCO dataset, possibly leading to a decrease in performance. It is also possible to retrain parts of the feature detector, allowing the network to learn new features from the dataset. To investigate which layers we can retrain without losing the benefits of transfer learning, we train a number of models with different parts of the feature detector frozen. We train a model starting from the 3rd, 4th and 5th ResNet stage and we perform an experiment where we retrain all layers. For this experiment, we only use the fully random synthetic dataset. Networks are again initialised with a model pre-trained on COCO and data augmentation is used.

\begin{table}
    \begin{center}
    \begin{tabular}{|l|c|}
    \hline
    Layers Retrained & AP \\
    \hline\hline
    All         & \textbf{81.26} \\
    Stage 3+   & 76.71 \\
    Stage 4+   & 80.77 \\
    Stage 5+   & 77.13 \\
    Heads       & 71.52 \\
    \hline
    \end{tabular}
    \end{center}
    \caption{Performance in AP on the real test set of models trained with transfer learning, but with different layers retrained.}
    \label{tab:retrain_layers}
\end{table}

The results for this experiment are shown in Table~\ref{tab:retrain_layers}. The best performing model is the model where all layers were retrained. The model where only the detection heads were retrained has the worst performance, falling at least five AP points below the other models. This shows us that while it is useful to initialise a network with transfer learning, it is important to let the network learn new features from the synthetic dataset as well. 

\subsection{Leveraging Real Images}
So far we have only considered using strictly synthetic images. It is however sometimes possible that some real images with their labels are available as well. In the following experiments we try to examine whether it is useful to use real images besides synthetic images and how to best use the real images. This can be done in many ways, the two most straightforward and widely used are: mixing the real images with the synthetic training images during training, and fine-tuning a network trained on synthetic images with real images afterwards. We will be testing both these methods. Furthermore, we analyse how many real images to use. We perform tests with different ratios of real to synthetic images, keeping the total amount of images constant at 3510. For these experiments we use the real and fully randomised datasets and train with transfer learning and data augmentation. For the initial training stage we only retrain the heads of the network. When fine-tuning, we retrain the entire network with a lower learning rate of $0.0001$.

\begin{figure}
    \centering
    \includegraphics[width=0.8\textwidth]{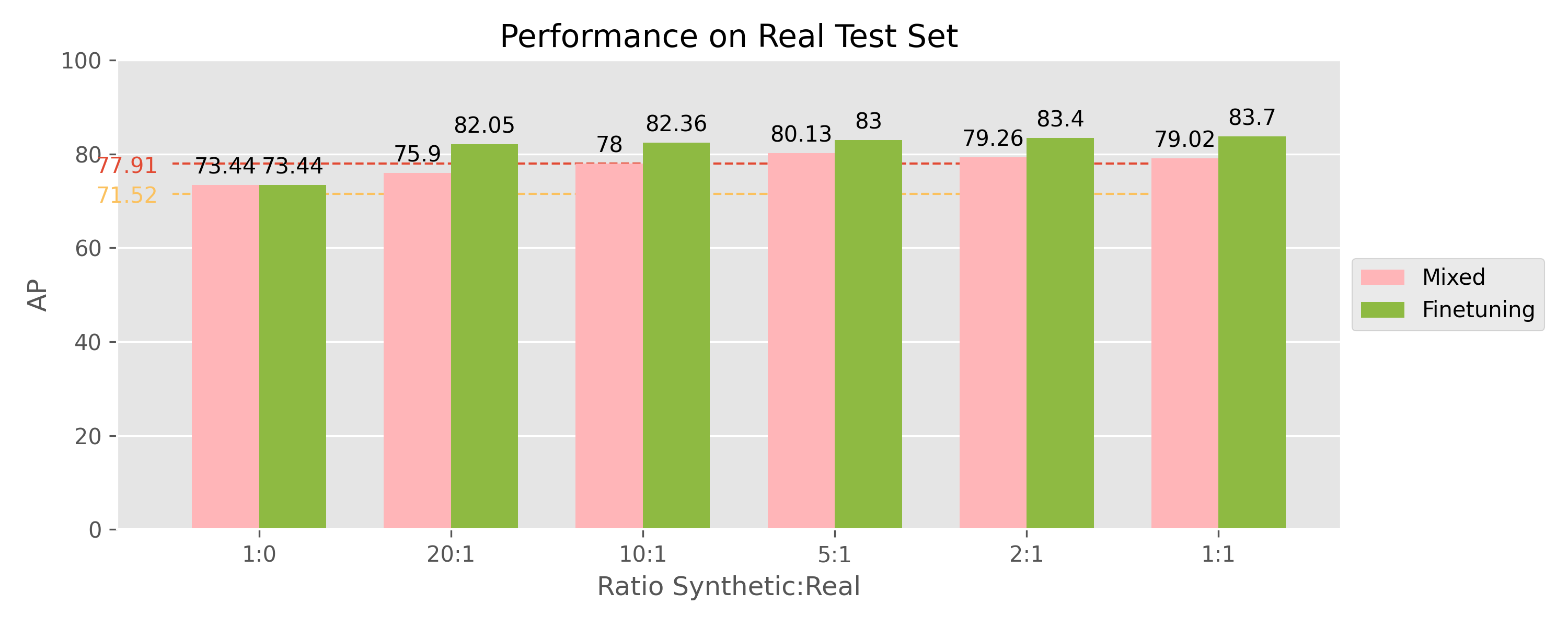}
    \caption{Performance in AP on the real test set of models trained with both synthetic and real data. The real data is either mixed in with the synthetic or used for fine tuning. The synthetic and real data is used in different ratios. The horizontal lines represent the performance of models trained on purely synthetic and real data, taken from previous experiments.}
    \label{fig:exp5}
\end{figure}

In Figure~\ref{fig:exp5} the results of the different ratios of synthetic to real data are shown for the different techniques. The horizontal lines represent the performance of the models trained on the real and fully random datasets. When mixing real images with the synthetic dataset we see that adding only a small amount already gives a performance boost. Adding a larger ratio of real images improves the performance even more, until the five to one ratio. The best performing model in this strategy achieves an AP of 80.13, which is an improvement over the model trained on real data. The models trained by fine-tuning on real data see a large benefit from this technique. Even when using only a small amount of real images the performance increases to 82.05 AP, this is 10 AP points above the model trained on the large random synthetic dataset and almost five AP points above the model trained on real data. Increasing the ratio of real images slightly increases the performance for the fine-tuning, reaching 83.7 AP for the one to one ratio.\\
From these results we can conclude that combining real and synthetic images can lead to an increase in performance compared to training on only one of the two. Additionally, we find that fine-tuning is the best way to use real images and that only a small amount of real images can already make a significant difference.

\section{Conclusion}
Throughout this paper we defined a number of experiments, training object detection models using different techniques on the DIMO dataset. We evaluated all these models, trained on synthetic data, on real data from the same problem domain. The goal was to acquire useful guidelines on how to generate data for deep learning and how to properly use this data. \\
Our experiments offer unique insights in how different variations of synthetic datasets perform on real data, using different training techniques. We show that modelling the lighting conditions and poses of a synthetic dataset to match the real target domain is beneficial towards generalisation, but only if transfer learning is used. When using transfer learning, we show that it is not beneficial to freeze the layers of the feature detector. It is better to retrain the entire network. This is contrary to some current research~\cite{pretrained_features}, so we argue this should be considered on a per-problem basis. Additionally, we investigated how to leverage real images. In our experiments we find that adding a small amount of real images is beneficial and that fine-tuning is the best method to do so. This is in line with the current state of the art~\cite{real_data}.\\
The Dataset of Industrial Metal Objects is a fairly simple dataset in terms of scene composition, as it contains no unknown objects and covers only a limited range of camera positions. Yet, it is the only dataset that includes the kind of controlled variations needed for these experiments and the scenes depicted in the dataset are highly relevant for industrial applications. We therefore believe that the recommendations made in this paper could serve as guidelines to generate data and train models for new problem domains. Future research should try to extrapolate these findings to different and more complex domains by generating other datasets with these controlled variations. 

\section*{Acknowledgements}
This study was supported by the Special Research Fund (BOF) of Hasselt University. The mandate ID is BOF20OWB24. Research was done in alignment with Flanders Make's PILS SBO project (R-9874).

\bibliography{references}

\begin{thebibliography}{21}
\providecommand{\natexlab}[1]{#1}
\providecommand{\url}[1]{\texttt{#1}}
\expandafter\ifx\csname urlstyle\endcsname\relax
  \providecommand{\doi}[1]{doi: #1}\else
  \providecommand{\doi}{doi: \begingroup \urlstyle{rm}\Url}\fi

\bibitem[Abdulla(2017)]{matterport_maskrcnn}
Waleed Abdulla.
\newblock Mask r-cnn for object detection and instance segmentation on keras
  and tensorflow.
\newblock \url{https://github.com/matterport/Mask_RCNN}, 2017.

\bibitem[Bai et~al.(2020)Bai, Li, Yang, Song, Li, and Zhang]{robot_control}
Qiang Bai, Shaobo Li, Jing Yang, Qisong Song, Zhiang Li, and Xingxing Zhang.
\newblock Object detection recognition and robot grasping based on machine
  learning: A survey.
\newblock \emph{IEEE Access}, 8:\penalty0 181855--181879, 2020.
\newblock \doi{10.1109/ACCESS.2020.3028740}.

\bibitem[De~Roovere et~al.(2022)De~Roovere, Moonen, Michiels, and
  wyffels]{DIMO}
Peter De~Roovere, Steven Moonen, Nick Michiels, and Francis wyffels.
\newblock Dataset of industrial metal objects, 2022.
\newblock URL \url{https://arxiv.org/abs/2208.04052}.

\bibitem[Gaidon et~al.(2016)Gaidon, Wang, Cabon, and Vig]{virtual_kitti}
Adrien Gaidon, Qiao Wang, Yohann Cabon, and Eleonora Vig.
\newblock Virtualworlds as proxy for multi-object tracking analysis.
\newblock In \emph{2016 IEEE Conference on Computer Vision and Pattern
  Recognition (CVPR)}, pages 4340--4349, 2016.
\newblock \doi{10.1109/CVPR.2016.470}.

\bibitem[Geiger et~al.(2012)Geiger, Lenz, and Urtasun]{kitti}
Andreas Geiger, Philip Lenz, and Raquel Urtasun.
\newblock Are we ready for autonomous driving? the kitti vision benchmark
  suite.
\newblock In \emph{Conference on Computer Vision and Pattern Recognition
  (CVPR)}, 2012.

\bibitem[He et~al.(2016)He, Zhang, Ren, and Sun]{resnet}
K.~He, X.~Zhang, S.~Ren, and J.~Sun.
\newblock Deep residual learning for image recognition.
\newblock In \emph{2016 IEEE Conference on Computer Vision and Pattern
  Recognition (CVPR)}, pages 770--778, Los Alamitos, CA, USA, jun 2016. IEEE
  Computer Society.
\newblock \doi{10.1109/CVPR.2016.90}.
\newblock URL \url{https://doi.ieeecomputersociety.org/10.1109/CVPR.2016.90}.

\bibitem[He et~al.(2017)He, Gkioxari, Dollár, and Girshick]{maskrcnn}
Kaiming He, Georgia Gkioxari, Piotr Dollár, and Ross Girshick.
\newblock Mask r-cnn.
\newblock In \emph{2017 IEEE International Conference on Computer Vision
  (ICCV)}, pages 2980--2988, 2017.
\newblock \doi{10.1109/ICCV.2017.322}.

\bibitem[Hinterstoisser et~al.(2019)Hinterstoisser, Lepetit, Wohlhart, and
  Konolige]{pretrained_features}
Stefan Hinterstoisser, Vincent Lepetit, Paul Wohlhart, and Kurt Konolige.
\newblock On pre-trained image features and synthetic images for deep learning.
\newblock In Laura Leal-Taix{\'e} and Stefan Roth, editors, \emph{Computer
  Vision -- ECCV 2018 Workshops}, pages 682--697, Cham, 2019. Springer
  International Publishing.
\newblock ISBN 978-3-030-11009-3.

\bibitem[Hodaň et~al.(2019)Hodaň, Vineet, Gal, Shalev, Hanzelka, Connell,
  Urbina, Sinha, and Guenter]{photorealism_hodan}
Tomáš Hodaň, Vibhav Vineet, Ran Gal, Emanuel Shalev, Jon Hanzelka, Treb
  Connell, Pedro Urbina, Sudipta~N. Sinha, and Brian Guenter.
\newblock Photorealistic image synthesis for object instance detection.
\newblock In \emph{2019 IEEE International Conference on Image Processing
  (ICIP)}, pages 66--70, 2019.
\newblock \doi{10.1109/ICIP.2019.8803821}.

\bibitem[Lin et~al.(2014)Lin, Maire, Belongie, Hays, Perona, Ramanan,
  Doll{\'a}r, and Zitnick]{coco_dataset}
Tsung-Yi Lin, Michael Maire, Serge Belongie, James Hays, Pietro Perona, Deva
  Ramanan, Piotr Doll{\'a}r, and C.~Lawrence Zitnick.
\newblock Microsoft coco: Common objects in context.
\newblock In David Fleet, Tomas Pajdla, Bernt Schiele, and Tinne Tuytelaars,
  editors, \emph{Computer Vision -- ECCV 2014}, pages 740--755, Cham, 2014.
  Springer International Publishing.
\newblock ISBN 978-3-319-10602-1.

\bibitem[Liu et~al.(2022)Liu, Mao, Wu, Feichtenhofer, Darrell, and
  Xie]{convnext}
Zhuang Liu, Hanzi Mao, Chao-Yuan Wu, Christoph Feichtenhofer, Trevor Darrell,
  and Saining Xie.
\newblock A convnet for the 2020s, 2022.
\newblock URL \url{https://arxiv.org/abs/2201.03545}.

\bibitem[Movshovitz-Attias et~al.(2016)Movshovitz-Attias, Kanade, and
  Sheikh]{photorealism_movshovitz}
Yair Movshovitz-Attias, Takeo Kanade, and Yaser Sheikh.
\newblock How useful is photo-realistic rendering for visual learning?
\newblock In Gang Hua and Herv{\'e} J{\'e}gou, editors, \emph{Computer Vision
  -- ECCV 2016 Workshops}, pages 202--217, Cham, 2016. Springer International
  Publishing.
\newblock ISBN 978-3-319-49409-8.

\bibitem[Northcutt et~al.(2021)Northcutt, Athalye, and Mueller]{dataset_error}
Curtis~G Northcutt, Anish Athalye, and Jonas Mueller.
\newblock Pervasive label errors in test sets destabilize machine learning
  benchmarks.
\newblock In \emph{Thirty-fifth Conference on Neural Information Processing
  Systems Datasets and Benchmarks Track (Round 1)}, 2021.
\newblock URL \url{https://openreview.net/forum?id=XccDXrDNLek}.

\bibitem[Nowruzi et~al.(2019)Nowruzi, Kapoor, Kolhatkar, Hassanat,
  Lagani{\`e}re, and Rebut]{real_data}
Farzan~Erlik Nowruzi, Prince Kapoor, Dhanvin Kolhatkar, Fahed~Al Hassanat,
  Robert Lagani{\`e}re, and Julien Rebut.
\newblock How much real data do we actually need: Analyzing object detection
  performance using synthetic and real data.
\newblock \emph{ArXiv}, abs/1907.07061, 2019.

\bibitem[Sreenu and Durai(2019)]{surveillance}
G.~Sreenu and M~A Durai.
\newblock Intelligent video surveillance: a review through deep learning
  techniques for crowd analysis.
\newblock \emph{Journal of Big Data}, 6:\penalty0 48, 06 2019.
\newblock \doi{10.1186/s40537-019-0212-5}.

\bibitem[Tobin et~al.(2017)Tobin, Fong, Ray, Schneider, Zaremba, and
  Abbeel]{domain_randomization_tobin}
Josh Tobin, Rachel Fong, Alex Ray, Jonas Schneider, Wojciech Zaremba, and
  Pieter Abbeel.
\newblock Domain randomization for transferring deep neural networks from
  simulation to the real world.
\newblock In \emph{2017 IEEE/RSJ International Conference on Intelligent Robots
  and Systems (IROS)}, pages 23--30, 2017.
\newblock \doi{10.1109/IROS.2017.8202133}.

\bibitem[Tommasi et~al.(2017)Tommasi, Patricia, Caputo, and
  Tuytelaars]{dataset_bias}
Tatiana Tommasi, Novi Patricia, Barbara Caputo, and Tinne Tuytelaars.
\newblock A deeper look at dataset bias.
\newblock In \emph{Domain adaptation in computer vision applications}, pages
  37--55. Springer, 2017.

\bibitem[Tremblay et~al.(2018)Tremblay, Prakash, Acuna, Brophy, Jampani, Anil,
  To, Cameracci, Boochoon, and Birchfield]{domain_randomization_tremblay}
Jonathan Tremblay, Aayush Prakash, David Acuna, Mark Brophy, Varun Jampani, Cem
  Anil, Thang To, Eric Cameracci, Shaad Boochoon, and Stan Birchfield.
\newblock Training deep networks with synthetic data: Bridging the reality gap
  by domain randomization.
\newblock In \emph{Proceedings of the IEEE conference on computer vision and
  pattern recognition workshops}, pages 969--977, 2018.

\bibitem[Wang et~al.(2022)Wang, Bochkovskiy, and Liao]{yolov7}
Chien-Yao Wang, Alexey Bochkovskiy, and Hong-Yuan~Mark Liao.
\newblock Yolov7: Trainable bag-of-freebies sets new state-of-the-art for
  real-time object detectors, 2022.
\newblock URL \url{https://arxiv.org/abs/2207.02696}.

\bibitem[Yang et~al.(2020)Yang, Li, Wang, Dong, Wang, and
  Tang]{product_inspection}
Jing Yang, Shaobo Li, Zheng Wang, Hao Dong, Jun Wang, and Shihao Tang.
\newblock Using deep learning to detect defects in manufacturing: A
  comprehensive survey and current challenges.
\newblock \emph{Materials}, 13\penalty0 (24), 2020.
\newblock ISSN 1996-1944.
\newblock \doi{10.3390/ma13245755}.
\newblock URL \url{https://www.mdpi.com/1996-1944/13/24/5755}.

\bibitem[Zhang et~al.(2022)Zhang, Li, Liu, Zhang, Su, Zhu, Ni, and Shum]{dino}
Hao Zhang, Feng Li, Shilong Liu, Lei Zhang, Hang Su, Jun Zhu, Lionel~M. Ni, and
  Heung-Yeung Shum.
\newblock Dino: Detr with improved denoising anchor boxes for end-to-end object
  detection, 2022.
\newblock URL \url{https://arxiv.org/abs/2203.03605}.

\end{thebibliography}

\end{document}